\documentclass[11pt,journal,a4paper,twoside,twocolumn]{IEEEtran}

\ifCLASSINFOpdf
  \usepackage[pdftex]{graphicx}
  \graphicspath{{../pdf/}{../jpeg/}}
  \DeclareGraphicsExtensions{.pdf,.jpeg,.png}
\else
  \usepackage[dvips]{graphicx}
  \graphicspath{{../eps/}}
  \DeclareGraphicsExtensions{.eps}
  \fi
\usepackage{amsmath,amsxtra,amssymb,amsthm,latexsym,amscd,amsfonts}
\usepackage[utf8]{vntex}
\ifCLASSINFOpdf
\usepackage[pdftex]{graphicx}
\DeclareGraphicsExtensions{.pdf,.jpeg,.png,.jpg}
\else
\usepackage[dvips]{graphicx}
\DeclareGraphicsExtensions{.eps}
\fi
\usepackage{array}
\usepackage{epstopdf}
\usepackage{anysize}
\usepackage{fancyhdr}
\usepackage{xcolor}

\newcommand{\fix}[1]{\textcolor{black}{#1}}

\marginsize{2.0cm}{2.0cm}{2.0cm}{2.0cm}
\usepackage{multicol}
\usepackage{balance}
\makeatletter
\def\ScaleIfNeeded{\ifdim\Gin@nat@width>\linewidth\linewidth\else\Gin@nat@width\fi}

\begin{document}

\pagestyle{fancy}
\fancyhf{}

\fancyhead[RO]{Dat Tran-Anh, Khanh Linh Tran, Hoai-Nam Vu}
\fancyhead[LE]{LICENSE PLATE RECOGNITION BASED ON MULTI-ANGLE VIEW MODEL}

\thispagestyle{empty}
\title{LICENSE PLATE RECOGNITION BASED ON MULTI-ANGLE VIEW MODEL}
\pagenumbering{gobble}
\fancypagestyle{firstpage}{%
  \lhead{*Left Header for just the first page**}
  \rhead{**Right Header for just the first page**}
}

\author{
\IEEEauthorblockN{
\textbf{Dat Tran-Anh}\IEEEauthorrefmark{1},
\textbf{Khanh Linh Tran}\IEEEauthorrefmark{2},
\textbf{Hoai-Nam Vu}\IEEEauthorrefmark{2}
} \\
\IEEEauthorblockA{\IEEEauthorrefmark{1}Thuyloi University}\\
\IEEEauthorblockA{\IEEEauthorrefmark{2}Posts and Telecommunications Institute of Technology}\\
\thanks{Tác giả liên hệ: Vũ Hoài Nam, email: namvh@ptit.edu.vn}
}
\maketitle


\begin{abstract}

In the realm of research, the detection/recognition of text within images/videos captured by cameras constitutes a highly challenging problem for researchers. Despite certain advancements achieving high accuracy, current methods still require substantial improvements to be applicable in practical scenarios. Diverging from text detection in images/videos, this paper addresses the issue of text detection within license plates by amalgamating multiple frames of distinct perspectives. For each viewpoint, the proposed method extracts descriptive features characterizing the text components of the license plate, specifically corner points and area. Concretely, we present three viewpoints: view-1, view-2, and view-3, to identify the nearest neighboring components facilitating the restoration of text components from the same license plate line based on estimations of similarity levels and distance metrics. Subsequently, we employ the CnOCR method for text recognition within license plates. Experimental results on the self-collected dataset (PTITPlates), comprising pairs of images in various scenarios, and the publicly available Stanford Cars Dataset, demonstrate the superiority of the proposed method over existing approaches.
	 
\end{abstract}

\begin{IEEEkeywords}
deep learning, license plate recognition and detection. 
\end{IEEEkeywords}
\IEEEpeerreviewmaketitle
\section{INTRODUCTION}

\begin{figure}[ht]
  \centering
  \includegraphics[width=0.45\textwidth]{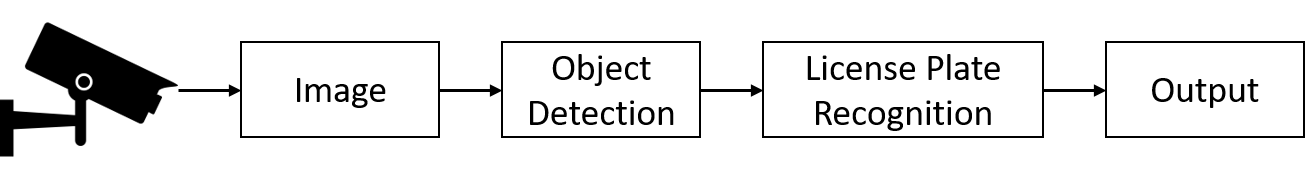}
  \caption{Image processing pipeline for License Plate Recognition.}
  \label{Fig:0}
\end{figure}

In recent decades, the traffic situation has become significantly more complex due to the global population increase \cite{Olayode}. Intelligent Transportation Systems (ITS) have been developed as a solution to the global traffic issue. To deploy ITS models, the management and automated recognition of vehicle license plates are considered crucial components. Figure \ref{Fig:0} illustrates the basic flow of license plate recognition software. License Plate Recognition (LPR) using smart camera systems typically involves four steps: firstly, the conversion of camera images into a format suitable for computer processing; next, the identification of regions of interest within the monitored camera image; subsequently, the recognition of characters on the license plate; finally, the presentation of the license plate recognition results \cite{Shashirangana}.

The traditional approach views a vehicle license plate as a region of interest and recognizes the characters as a sequence, followed by comparing two sequences to identify the vehicle. In the research study by Vaishnav et al. \cite{Vaishnav}, the authors proposed a system that utilizes optical character recognition techniques and compares characters with stored templates to obtain specific information about the license plate. Comparing license plate numbers yields accurate results; however, this method is effective only when the license plate is clearly displayed in the image. If the license plate is obscured or not securely attached to the registered vehicle, inaccurate results may be obtained.

These issues can be addressed by utilizing additional features of the license plate for comparison. Unlike the traditional approach, deep learning models based on multi-layered architectures can learn license plate characteristics at different levels. These deep learning models take raw images (without feature extraction) as direct input. Most deep learning methods for license plate recognition learn the plate features within the model \cite{Lu, Slimani}. Kessentini et al. \cite{Kessentini} designed a two-stream architecture: (i) stream 1 processes input vehicle features; (ii) stream 2 processes input license plate features. However, this method only performs license plate recognition from a single viewing angle. Recently, with the rapid development of widely distributed camera systems, multi-angle data collection has become feasible. Consequently, license plate recognition systems can benefit from this multi-angle data. Different viewing angles of the license plate provide the opportunity to extract diverse features, which are useful for recognition. In this study, we apply the YOLOv8 architecture, take license plates from multiple viewing angles as input, and propose a deep learning model for accurate license plate recognition in various real-world situations.

\section{RELATED WORK}
\label{Sec:NghienCuuLienQuan}

License plate recognition is divided into two main stages: (1) the license plate detection stage and (2) the license plate recognition stage.

\subsection{License plate detection stage.}

Recently, computer vision-based methods for license plate detection have garnered significant attention in Intelligent Transportation System (ITS) applications. Achieving highly accurate license plate detection is a fundamental component of traffic monitoring aimed at increasing safety and automation \cite{Ma}. A comprehensive survey evaluating license plate detection systems is presented in \cite{Burkpalli}. With the recent strong advancements of deep learning algorithms in various image processing and pattern recognition domains \cite{Abiodun, Zhang, Bernardo, Caggiano}, single-camera object detection systems based on Convolutional Neural Networks (CNNs) have been investigated \cite{Kattenborn, Lu}. However, these single-camera systems might not be able to detect partially obscured license plates in congested traffic contexts. An alternative approach to overcome this challenge is to employ multi-camera systems and integrate information from each independent camera stream \cite{Tourani, Izidio}. Mukhija et al. \cite{Mukhija} proposed a method based on wavelet transform and Empirical Mode Decomposition (EMD) for license plate localization in images, addressing real-world challenges such as lighting variations, complex backgrounds, and changes in surroundings. MASK-RCNN \cite{Vuola} introduced a simplified, flexible, and popular segmentation framework that can create masks for potential objects and accurately segment targets. The YOLO model and its upgraded versions \cite{Isa} consider object detection as a regression task, enabling efficient object detection with high accuracy and fast speed. Deep learning models and architectures based on YOLO are increasingly popular in the research community. Therefore, YOLOv8 is utilized in this paper as the framework for the license plate detection component in our system.

\subsection{License plate recognition stage.}

Some license plate recognition systems are designed to segment characters before recognizing them. Segmentation methods can be categorized into connected-component analysis \cite{Qiao}, projection profile analysis \cite{Imran}, prior character knowledge \cite{Zhou}, contour analysis around characters \cite{Pereira}, and combinations of these methods \cite{Gao}. It becomes evident that accurately classifying all characters within a license plate is challenging when the character segmentation component performs poorly. Consequently, some researchers focus on proposing reliable character segmentation methods for license plate recognition. Meanwhile, other studies concentrate on suggesting license plate recognition methods without character segmentation, transforming the problem into a sequence labeling task \cite{LinShao}. Leveraging the strengths of improved YOLO models, license plate characters have been segmented and recognized in \cite{Upadhyay}. The accuracy of character segmentation depends on the segmentation performance and can be affected by external conditions like light intensity, blurriness of the license plate, etc. These conditions can reduce the accuracy of license plate recognition. Currently, most researchers apply non-character segmentation methods. RPnet, proposed by Xu et al. \cite{XuJiang}, swiftly and accurately predicts license plate bounding boxes, simultaneously determining the corresponding license plate by extracting features of Regions of Interest (ROIs) from different convolutional layers. This model surpasses existing object detection and recognition models in both accuracy and speed, achieving a 98.5\% accuracy rate. 

\subsection{Character Recognition methods}

In order to recognize characters within images, many research groups \cite{NguyenJatowt} have relied on image features for identification. The CRNN study \cite{Jiao} initially combines CNN and RNN to extract sequential visual features from a specific text image. These features are then fed directly into the CTC decoding mechanism to predict the best character type for each time step. CTC \cite{Rais-Zadeh} in this context is a loss function used to train deep learning models. Most methods recognize characters in a unidirectional manner. For example, Lee et al. \cite{LyuYang} encoded input text images horizontally into 2D sequential image features, which are subsequently input into the corresponding regions with semantic information assistance from the previous time step. To mitigate mischaracterizations due to scene distortion and skewed distribution, Yang et al. \cite{Rais-Yang} introduced an improved module prior to character recognition. This module employed a spatial transformation network with multiple control point pairs. In our research framework, the CnOCR module is utilized, applying unidirectional character recognition to accurately locate character features and enhance the recognition performance of the model.

\section{PROPOSED METHOD}
\label{Sec:PhuongPhapDeXuat}

\begin{figure*}[ht]
  \centering
  \includegraphics[width=0.9\textwidth,height=8.3cm]{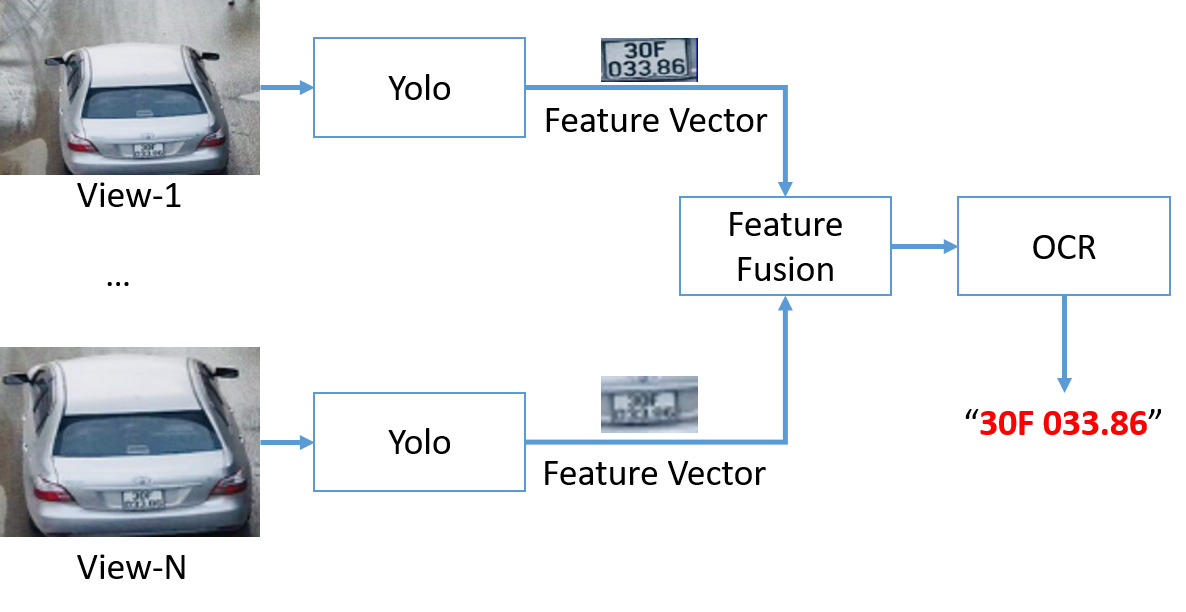}
  \caption{Overview of our proposed method.}
  \label{Fig:1}
\end{figure*}

Our proposed approach is depicted in Figure~\ref{Fig:1}, comprising three main components: (i) the YOLO model for license plate detection; (ii) the License Plate Image Fusion algorithm for selecting the highest-quality license plate image; and (iii) the OCR model for character feature extraction and license plate recognition. The input consists of individual frames captured from cameras, which are divided into different viewpoints (views) for each camera. Each viewpoint becomes the input for a YOLO model. In this paper, we optimize the YOLO license plate detection model based on experimental results from various benchmark datasets. The quality of license plate images varies across different views, including factors like angle, visibility, blurriness, and distortion. Thus, we develop a License Plate Image Fusion algorithm to combine similar license plate images into a single image with enhanced information. Finally, the fused license plate image is passed through the OCR model for license plate recognition.

\subsection{YOLO model}

The YOLOv8 model is employed to detect license plates appearing within frames. YOLOv8 is chosen due to its high accuracy in detection and fast processing times, making it suitable for real-time applications. Moreover, the YOLOv8 model provides various versions with different sizes, allowing deployment in diverse environments.

Our system processes high-resolution input images (3840 × 2160) decoded from high-resolution videos. We collected images of various vehicle types and real Vietnamese license plates to create a custom dataset. Subsequently, this custom dataset was used to train the YOLOv8 model in order to construct two custom detection pipelines. The detection model is capable of identifying seven different object types (six vehicle types and different Vietnamese license plates) within the input images. In this phase, vehicle types and license plate occurrences are detected by the detection model. If a license plate is detected, the license plate image is cropped and passed on to phase 2. In summary, the detection model is first called to identify vehicle types and license plates. In cases where the input image contains a large number of license plates, the iterative process of recognizing each license plate may take longer compared to scenarios with only one license plate present in the frame.

\subsection{Image Fusion algorithm}

In evaluating object detection methods, the Intersection over Union (IoU) metric \cite{ZhengWang} is commonly utilized. The IoU is a crucial measure for assessing the accuracy of object detection results. The underlying principle of the IoU is depicted in Figure 3. The IoU is calculated by dividing the area of intersection between the predicted bounding box and the ground-truth bounding box by the area of their union. This provides a quantitative measure of how well the predicted bounding box aligns with the actual object location. To enhance the IoU metric, the Generalized IoU (GIoU) was introduced to address situations where the IoU loss becomes zero when the predicted bounding box and the ground-truth bounding box do not directly overlap. Moreover, DIoU \cite{ZhengWang}, introduces the Euclidean distance between the center points of the predicted and ground-truth bounding boxes based on the GIoU metric. This addition further refines the evaluation by considering the spatial distance between the bounding boxes, thereby aiding in speeding up the convergence of object detection model training.

\begin{figure}[h]
  \centering
  \includegraphics[width=0.5\textwidth,height=5.7cm]{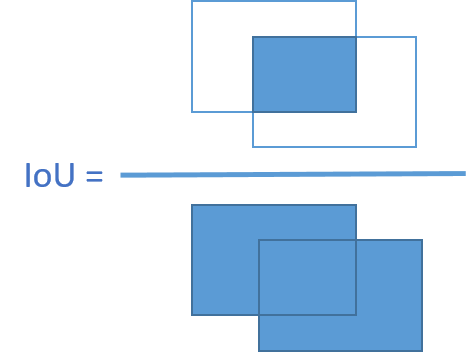}
  \caption{IoU score.}
  \label{Fig:2}
\end{figure}

\begin{figure}[h]
  \centering
  \includegraphics[width=0.5\textwidth]{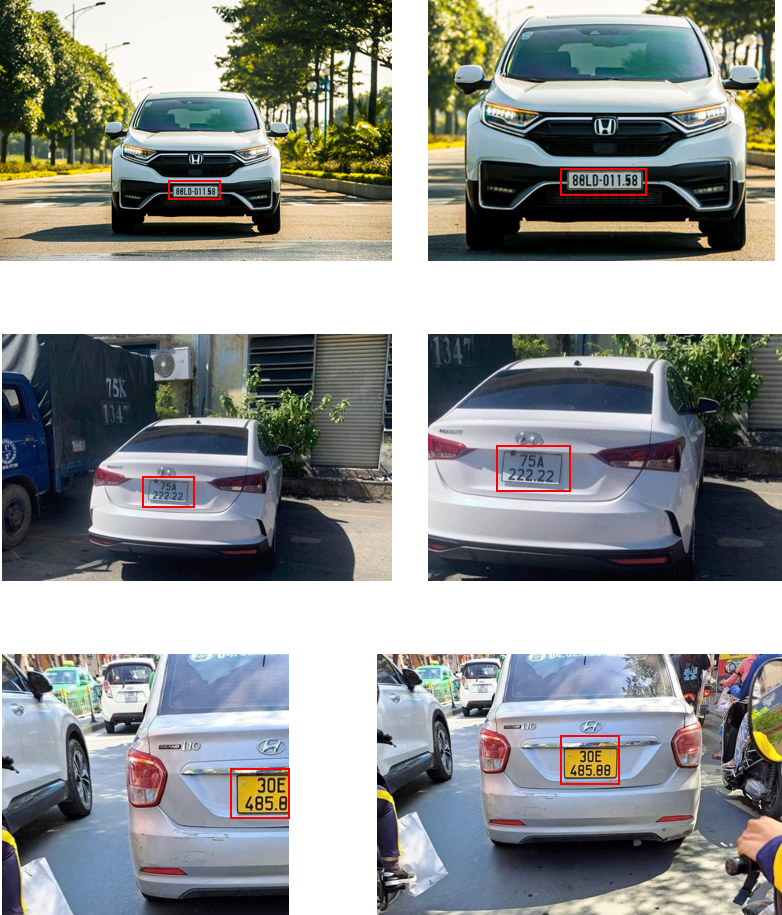}
  \caption{Some examples of license plate detection}
  \label{Fig:22}
\end{figure}

While the IoU metric has undergone multiple refinements across its various iterations, it still may not be inherently suitable for the construction of automated license plate recognition (LPR) models.  This stems from the fact that within LPR models, if a license plate is detected with an excess of supplementary information (the detection area surpasses the actual license plate area), subsequent license plate recognition models may encounter fewer hindrances in accurately extracting characters from the detected license plate.  However, when a license plate is detected with information deficits (the detection area is smaller than the genuine license plate area), this can pose challenges for subsequent recognition modules, thereby resulting in recognition inaccuracies. To expound upon this matter, we offer illustrative examples as depicted in Figure \ref{Fig:22}. It is evident that the predicted bounding box, as depicted in Figure \ref{Fig:22}, manifests a larger area than the actual license plate region.

However, if the area of the predicted bounding box happens to be smaller than the actual license plate area, it can result in certain characters not being encompassed within the predicted region. Such information loss can lead to errors in the final license plate recognition outcome. Consequently, models based on loss functions utilizing the existing IoU metric are unable to effectively address these issues, as they assign similar priorities to regions with both information loss and surplus. A novel loss function based on the IoU metric is imperative to achieve a more balanced treatment between information loss and surplus, thereby providing a more effective means of handling these challenges.

After the IoU calculation, two regions are identified: (1) Non-overlapping region; (2) Overlapping region. For the overlapping region, we employ an image fusion method to generate an image with the best quality. Given the source images denoted as $\mathbf{I_1}$ and $\mathbf{I_2}$, a DenseNet model is trained to produce the fused image. The output of the feature extractor comprises feature maps $\phi C_1 (\mathbf{I_1}),... , \phi C_5 (\mathbf{I_1})$ and $\phi C_1 (\mathbf{I_2}),... , \phi C_5 (\mathbf{I_2})$ \fix{where $C_i$ represents a specific layer within the feature extractor and $\phi$ is the feature extractor}. Subsequently, an information measure is performed on these feature maps, resulting in two measures denoted as $g\mathbf{I_1}$ and $g\mathbf{I_2}$. In the subsequent processing, the degree of information preservation is denoted as $\omega_1$ and $\omega_2$. $\mathbf{I_1}$, $\mathbf{I_2}$, $\mathbf{I_f}$, $\omega_1$, and $\omega_2$ are utilized in the loss function without requiring ground truth labels. During the training phase, $\omega_1$ and $\omega_2$ are computed to determine the loss function. Afterwards, a DenseNet module is optimized to minimize the loss value. In the testing phase, $\omega_1$ and $\omega_2$ need not be computed as the DenseNet model has been optimized. Therefore, $\omega_1$ and $\omega_2$ are defined as:

\begin{equation}
[\omega_1,\omega_1]=softmax([\frac{g\mathbf{I_1}}{c}, \frac{g\mathbf{I_1}}{c}])
\end{equation}

In this context, we employ the softmax function to map $\frac{g\mathbf{I_1}}{c}$ and $\frac{g\mathbf{I_2}}{c}$ to real numbers within the range of 0 to 1, ensuring that the sum of $\omega_1$ and $\omega_2$ equals 1. Subsequently, $\omega_1$ and $\omega_2$ are utilized in the loss function to control the degree of information preservation of specific source images.

The loss function is primarily designed to preserve essential information and to train a single model that can be applied to various tasks. It is defined as follows:

\begin{equation}
\mathcal{L} = \mathbb{E}(\omega_1 \cdot MSE_{\mathbf{I_f}, \mathbf{I_1}} + \omega_2 \cdot MSE_{\mathbf{I_f}, \mathbf{I_2}})
\end{equation}

This loss function is then utilized to train the feature aggregation model, which combines features from multiple different frames into an optimized feature representation for the license plate character recognition task in images.
\subsection{OCR model}

\begin{figure*}[h]
  \centering
  \includegraphics[width=0.97\textwidth]{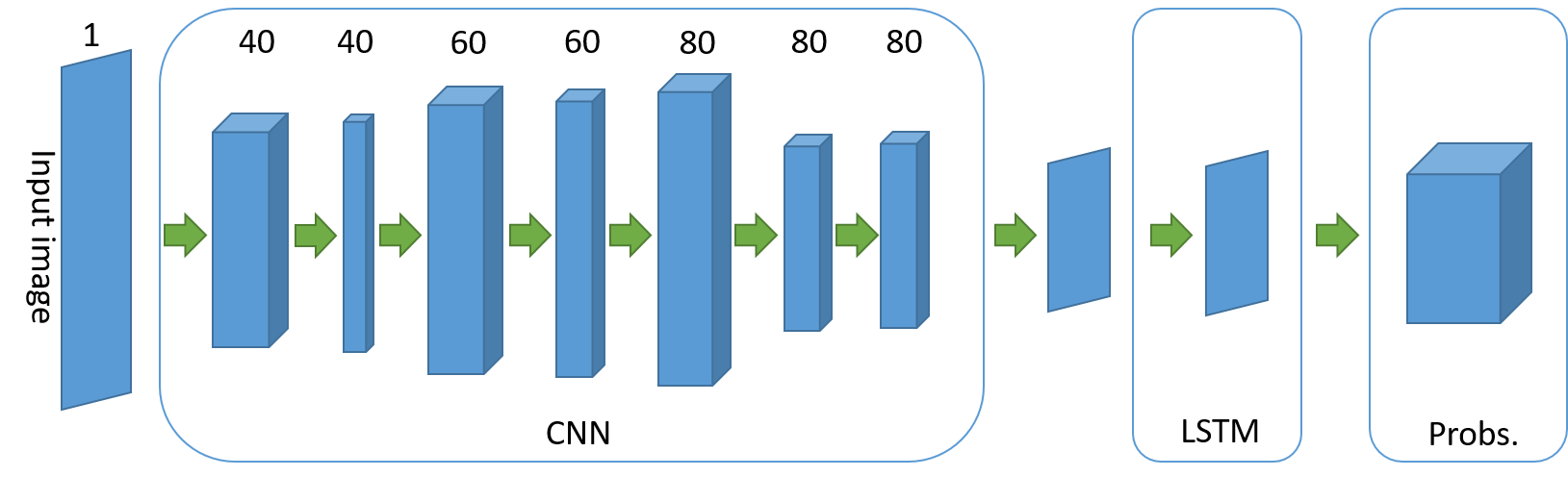}
  \caption{OCR model.}
  \label{Fig:5}
\end{figure*}

Figure \ref{Fig:5} illustrates the chosen architecture of the CnOCR network for the character recognition part. Initially, a convolutional layer with 40 kernels of size 3 × 3 is applied to the input image, which is a matrix block, to extract basic features. A subsequent pooling layer aims to reduce the resolution by selecting the most prominent features within a 1 × 2 region. Two additional convolutional sets (60 and 80 kernels) and max pooling layers are added. However, the final pooling layer employs a filter size of 2 × 2. Traditional architectures usually perform 2 × 2 pooling, halving both dimensions. In contrast, we apply two pooling layers to halve only the height, not the width. The reason is that the maximum number of predicted labels corresponds to the size of the time axis of the last layer, which is the width in our case. Due to the dense and overlapping nature of some license plates characters, we incorporate only a 2 × 2 pooling layer. After the final pooling layer, a fourth convolutional layer is added with an 80-sized kernel, followed by a bidirectional LSTM layer with 100 hidden nodes.\fix{The LSTM layer is instrumental in capturing contextual information and dependencies between characters within the text while convolutional layers are used as feature extractors to analyze the visual characteristics of characters and text regions.} Lastly, a dense layer with softmax activation transforms the 100 output nodes at each position into probabilities for the 53 (52 + blank) target characters at each horizontal position for character recognition. The key advantage of CnOCR lies in its fast prediction speed, achieving both accuracy and prediction time of 0.03 seconds for a single license plate.

\subsection{The license plate dataset and configurations}

The PTITPlates dataset consists of 500 license plate images labeled using the LabelMe tool. We collected these images through cameras placed in various industrial and road areas. The images capture different angles and have been filtered to include only those with visible license plates, which aids in training and testing the proposed model. The training parameters for our proposed method are presented in Table~\ref{tab:t1}. The total trainable parameters of the YOLOv8 model are around 11 million, while for the feature fusion and OCR models, the parameter count is smaller. The optimization algorithm we employ is Stochastic Gradient Descent, and the loss function used is cross-entropy. 

\begin{table}[]
\centering
\caption{Parameters of the proposed model:}
\label{tab:t1}
\begin{tabular}{|lll|}
\hline
\multicolumn{1}{|c|}{\textbf{Parameters}} & \multicolumn{1}{c|}{\textbf{Layers}} & \multicolumn{1}{c|}{\textbf{Shapes}} \\ \hline
\multicolumn{1}{|l|}{928}             & \multicolumn{1}{l|}{Conv}            & {[}3, 32, 3, 2{]}                        \\ \hline
\multicolumn{1}{|l|}{18560}           & \multicolumn{1}{l|}{Conv}            & {[}32, 64, 3, 2{]}                       \\ \hline
\multicolumn{1}{|l|}{29056}           & \multicolumn{1}{l|}{C2f}             & {[}64, 64, 1, True{]}                    \\ \hline
\multicolumn{1}{|l|}{73984}           & \multicolumn{1}{l|}{Conv}            & {[}64, 128, 3, 2{]}                      \\ \hline
\multicolumn{1}{|l|}{197632}          & \multicolumn{1}{l|}{C2f}             & {[}128, 128, 2, True{]}                  \\ \hline
\multicolumn{1}{|l|}{295424}          & \multicolumn{1}{l|}{Conv}            & {[}128, 256, 3, 2{]}                     \\ \hline
\multicolumn{1}{|l|}{788480}          & \multicolumn{1}{l|}{C2f}             & {[}256, 256, 2, True{]}                  \\ \hline
\multicolumn{1}{|l|}{1180672}         & \multicolumn{1}{l|}{Conv}            & {[}256, 512, 3, 2{]}                     \\ \hline
\multicolumn{1}{|l|}{1838080}         & \multicolumn{1}{l|}{C2f}             & {[}512, 512, 1, True{]}                  \\ \hline
\multicolumn{1}{|l|}{656896}          & \multicolumn{1}{l|}{SPPF}            & {[}512, 512, 5{]}                        \\ \hline
\multicolumn{1}{|l|}{0}               & \multicolumn{1}{l|}{Upsample}        & {[}None, 2, 'nearest'{]}                 \\ \hline
\multicolumn{1}{|l|}{0}               & \multicolumn{1}{l|}{Concat}          & {[}1{]}                                  \\ \hline
\multicolumn{1}{|l|}{591360}          & \multicolumn{1}{l|}{C2f}             & {[}768, 256, 1{]}                        \\ \hline
\multicolumn{1}{|l|}{0}               & \multicolumn{1}{l|}{Upsample}        & {[}None, 2, 'nearest'{]}                 \\ \hline
\multicolumn{1}{|l|}{0}               & \multicolumn{1}{l|}{Concat}          & {[}1{]}                                  \\ \hline
\multicolumn{1}{|l|}{148224}          & \multicolumn{1}{l|}{C2f}             & {[}384, 128, 1{]}                        \\ \hline
\multicolumn{1}{|l|}{147712}          & \multicolumn{1}{l|}{Conv}            & {[}128, 128, 3, 2{]}                     \\ \hline
\multicolumn{1}{|l|}{0}               & \multicolumn{1}{l|}{Concat}          & {[}1{]}                                  \\ \hline
\multicolumn{1}{|l|}{493056}          & \multicolumn{1}{l|}{C2f}             & {[}384, 256, 1{]}                        \\ \hline
\multicolumn{1}{|l|}{590336}          & \multicolumn{1}{l|}{Conv}            & {[}256, 256, 3, 2{]}                     \\ \hline
\multicolumn{1}{|l|}{0}               & \multicolumn{1}{l|}{Concat}          & {[}1{]}                                  \\ \hline
\multicolumn{1}{|l|}{1969152}         & \multicolumn{1}{l|}{C2f}             & {[}768, 512, 1{]}                        \\ \hline
\multicolumn{1}{|l|}{2116435}         & \multicolumn{1}{l|}{Detect}          & {[}1, {[}128, 256, 512{]}{]}             \\ \hline
\multicolumn{3}{|l|}{\textbf{Model summary: 225 layers, 11135987 parameters}}                                           \\ \hline
\end{tabular}
\end{table}

\begin{figure*}[h]
  \centering
  \includegraphics[width=0.9\textwidth]{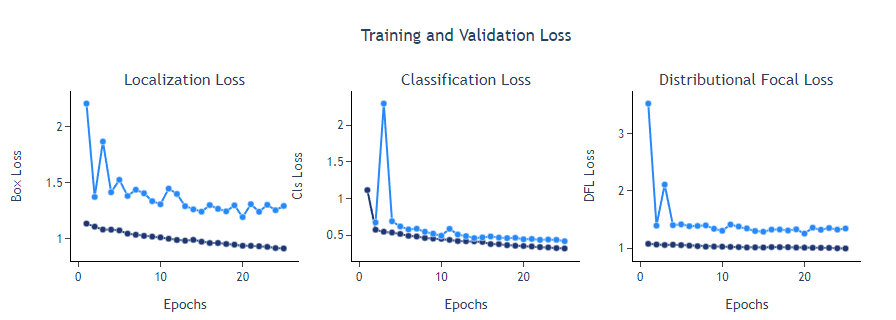}
  \caption{Transformation of the loss function during the training process}
  \label{Fig:6}
\end{figure*}

We utilize loss value transformation graphs over time to evaluate and compare the performance of different loss functions. These graphs provide a visual representation that allows us to observe and analyze the variation of the loss function during training. We present three graph images in Figure \ref{Fig:6}, corresponding to three different loss functions: (i) Localization Loss, depicting the training process of license plate detection; (ii) Classification Loss, illustrating the training of license plate classification and image fusion model; and (iii) Connectionist Temporal Classification Loss, indicating the character recognition process within the license plate.

Specifically, for the YOLOv8 loss function, the Variational Focal Loss (VFL) is employed as the classification loss, while the combination of Distribution Focal Loss (DFL) and Complete Intersection over Union Loss (CIOU) serves as the segmentation loss.
Similarly to conventional classification tasks, in the license plate classification step, we employ the Categorical Cross-Entropy loss with the following formula (equation \ref{CEL}):
\begin{equation}
    \mathcal{L}_{CE} = -\sum \frac{1}{N}(y_{true} * log(y_{pred}))
    \label{CEL}
\end{equation}
In which $y_{true}$ represents the ground truth vector (one-hot encoding) of the sample with a size of $C$, which is the number of different classes,
$y_{pred}$ is the predicted vector of the sample and $N$ is the number of data samples.

Finally, in the recognition model, we utilize the loss function with the general formula of CTC as follows:
\begin{equation}
    \mathcal{L}_{CTC} = -log(P(Y | X))
\end{equation}

Where $P(Y | X)$ represents the probability of the actual output sequence $Y$ given the input sequence $X$. 

The results of the training process are depicted in Figure~\ref{Fig:6}. The loss values and accuracy during the training process exhibit a pronounced fluctuation in the initial epochs, gradually stabilizing in the subsequent epochs, indicates that the model is learning and improving over time. \fix{If the loss values and accuracy exhibit instability during training then proposed model may not be suitable for the dataset. Our proposed model tends to converge to an optimal value after around 20 epochs. The training process was concluded after 25 epochs. The validation loss curves are generally close to the training loss curves,implying that the model is not overfitting to the training data. Additionally, the gradual, sustained  observed in the validation loss curves signifies a continuous enhancement in the model's capability to generalize and excel when applied to unobserved data.}

\begin{figure}[h]
  \centering
  \includegraphics[width=0.45\textwidth]{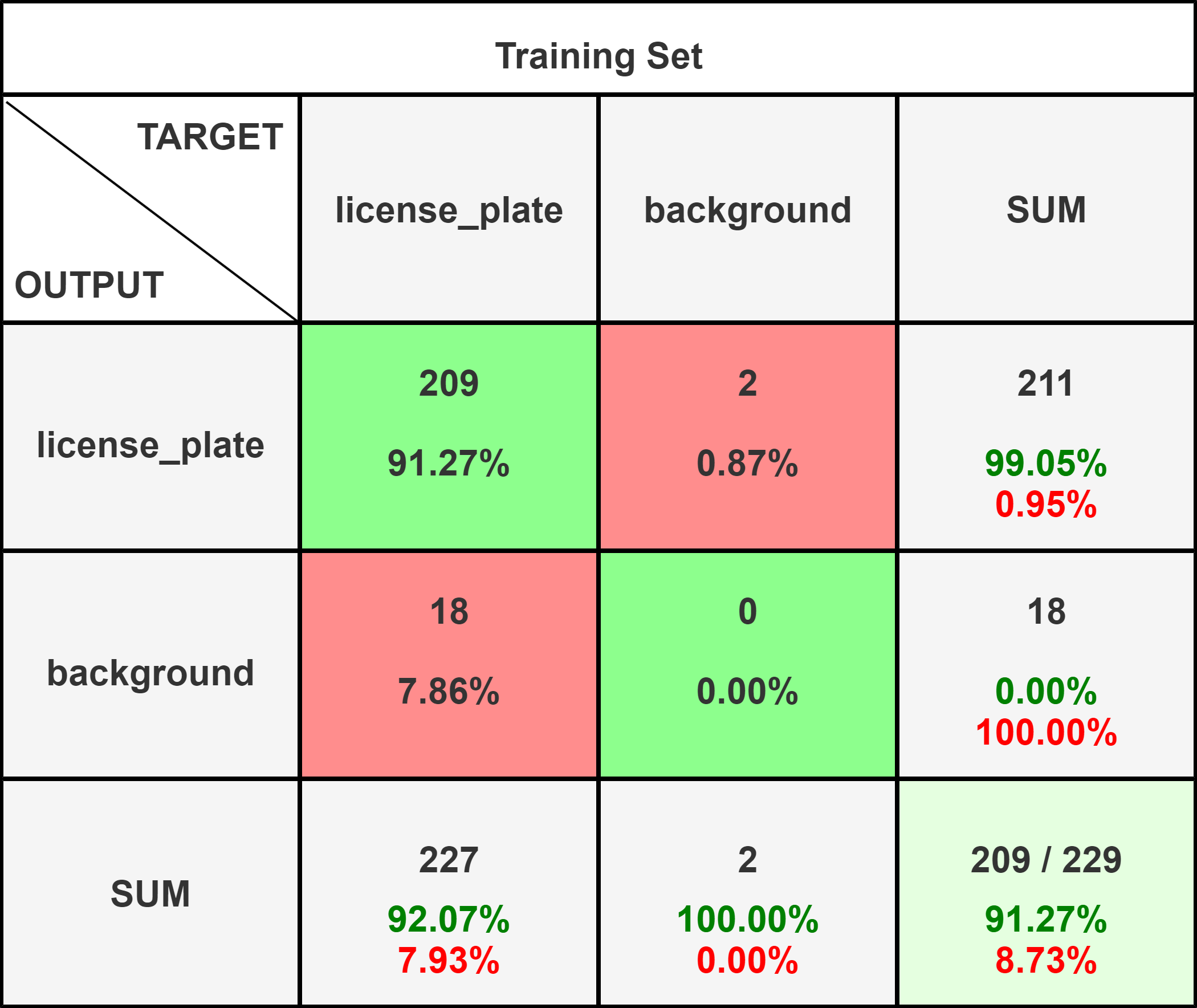}
  \caption{Confusion matrix of the proposed method.}
  \label{Fig:7}
\end{figure}

\section{EXPERIMENTAL RESULTS}
\subsection{Accuracy Evaluation}

We evaluated the models using the F1 score, considering equal importance in the accurate classification of each class. From the results presented in Table \ref{tab:t3}, it can be observed that our proposed method achieves higher F1 scores compared to the baseline methods, such as the YOLOv5 network which is currently considered one of the best methods, as well as other methods like YOLOv8 with Tesseract and YOLOv8 with CnOCR. This performance difference can be attributed to the variations in the angles of license plates and the impact of weather conditions. Our model effectively identifies the partitioning mechanism using blocks and image fusion, resulting in improved performance. The F1 score for the YOLOv5 network is only 75.2\% with the PTITPlates dataset and 78.6\% with the Stanford Cars dataset. In contrast, our proposed method achieves F1 scores of 91.3\% with the PTITPlates dataset and 90.8\% with the Stanford Cars dataset. \fix{This phenomenon arises due to the presence of substantial noise within the Stanford Cars dataset and its limited variety of viewing angles, as compared to the PTITPlates dataset. This observation underscores a vulnerability commonly associated with datasets that lack diversity in viewing angles.} The detailed confusion matrix of the proposed license plate detection model is provided in Figure \ref{Fig:7}. \fix{As displayed in the confusion matrix, the proposed method accurately categorized the majority of specific license plates, as evidenced by elevated counts of True Positives (TP). However, it encounters challenges in certain instances, particularly when confronted with blurred plates, leading to the presence of False Positives (FP) and False Negatives (FN).}

\begin{table}[h]
\centering
\caption{comparision results}
\label{tab:t3}
\begin{tabular}{|c|l|l|}
\hline
\textbf{Dataset}                                          & \multicolumn{1}{c|}{\textbf{Method}} & \multicolumn{1}{c|}{\textbf{F1-Score}} \\ \hline
{Stanford Cars}                            & YOLOv5 + Base OCR                    & 78.6                                   \\ \cline{2-3} 
                                                          & YOLOv8 + Tesseract OCR               & 80.3                                   \\ \cline{2-3} 
                                                          & YOLOv8 + CnOCR                       & 86.3                                   \\ \cline{2-3} 
                                                          & \textbf{Our proposed method}         & \textbf{90.8}                          \\ \hline
\multicolumn{1}{|l|}{{PTITPlates}} & YOLOv5 + Base OCR                    & 75.2                                   \\ \cline{2-3} 
\multicolumn{1}{|l|}{}                                    & YOLOv8 + Tesseract OCR               & 82.9                                   \\ \cline{2-3} 
\multicolumn{1}{|l|}{}                                    & YOLOv8 + CnOCR                       & 85.2                                   \\ \cline{2-3} 
\multicolumn{1}{|l|}{}                                    & \textbf{Our proposed method}         & \textbf{91.3}                          \\ \hline
\end{tabular}
\end{table}

\subsection{Real-world testing}

\begin{figure*}[ht]
  \centering
  \includegraphics[width=0.9\textwidth]{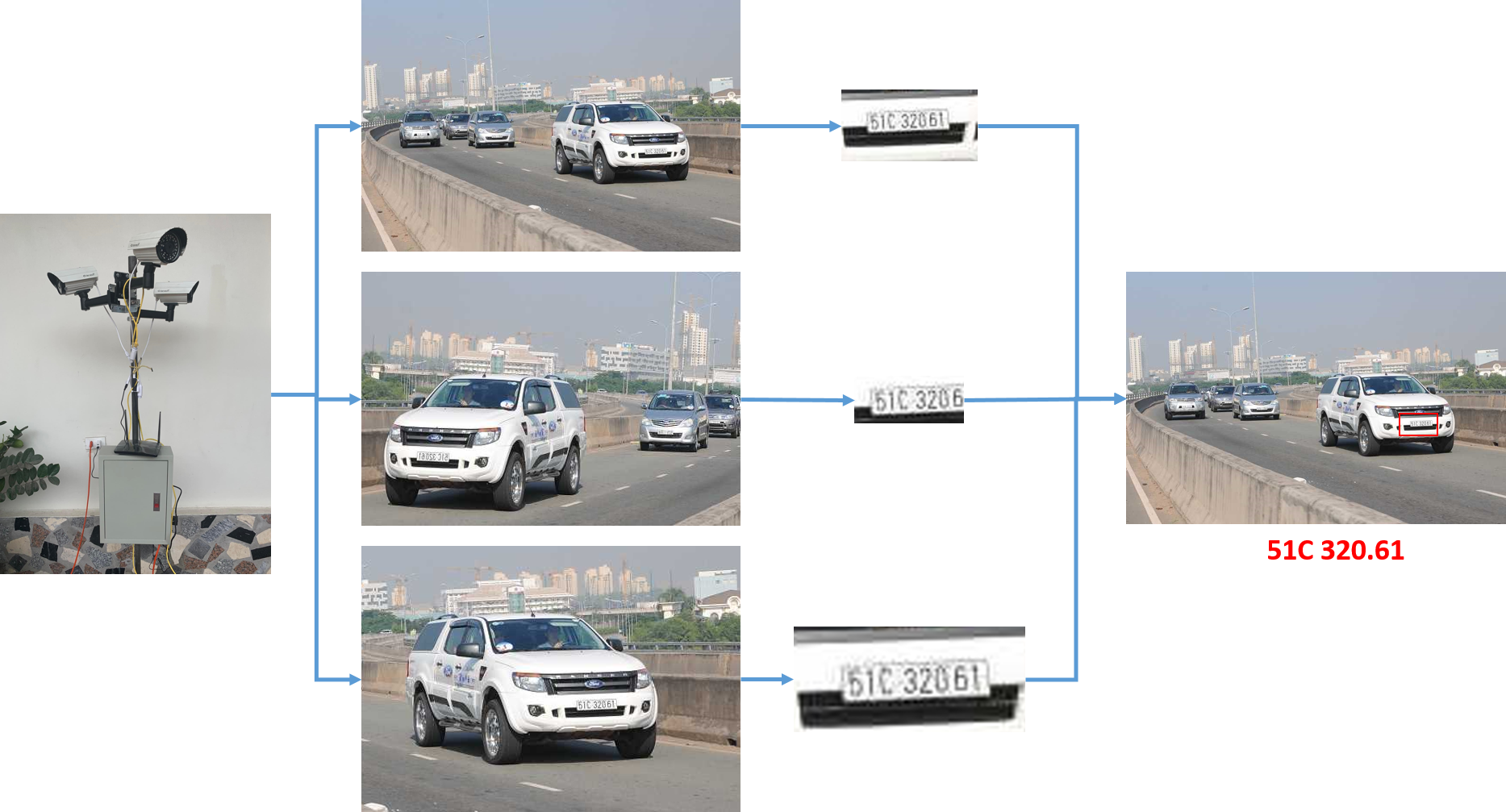}
  \caption{Some examples of predictions made by our model}
  \label{Fig:33}
\end{figure*}

In this experiment, we also integrated the model into a practical application. Some images of our model's predictions are shown in Figure \ref{Fig:33}. We deployed the system to handle 30 cameras, grouped into 10 columns, to monitor vehicle entries and exits in an industrial zone. Due to the substantial volume of incoming images, we designed a distributed system with 10 integrated API endpoints, each utilizing our model. On average, each model processed data from 3 cameras, achieving a license plate detection latency of 0.1 seconds. This result is quite impressive for the deployment of a license plate detection and recognition system.

\section{CONCLUSION}

In this paper, we proposed a feature fusion model from multiple perspectives based on the YOLOv8 and CnOCR architectures for license plate recognition. Through evaluations on various datasets, experimental results demonstrated that the proposed model achieved impressive results, with an F1 score of 91.3\% on the self-collected PTITPlates dataset. Notably, this dataset is noisy and affected by weather conditions, resulting in poor image quality of license plates. The proposed model showcased robust performance across diverse contexts with high accuracy. In the future, Generative Adversarial Networks (GANs) could be employed for data augmentation, simulating license plates from different regions to address class imbalance issues among Vietnamese character groups within the dataset. This could potentially enhance accuracy. Additionally, self-supervised learning models, such as zero-shot learning, could be utilized to fine-tune the network based on the localization accuracy of specific character features within license plates, potentially leading to even better results.

\bibliographystyle{IEEEtran}

\begin{thebibliography}{00}
\bibitem{Olayode} I. O. Olayode, L. K. Tartibu, M. O. Okwu, and U. F. Uchechi, “Intelligent transportation systems, un-signalized road intersections and traffic congestion in Johannesburg: A systematic review,” Procedia CIRP, vol. 91, no. March, pp. 844–850, 2020, doi: 10.1016/j.procir.2020.04.137.
\bibitem{Shashirangana} J. Shashirangana, H. Padmasiri, D. Meedeniya and C. Perera, "Automated License Plate Recognition: A Survey on Methods and Techniques," in IEEE Access, vol. 9, pp. 11203-11225, 2021, doi: 10.1109/ACCESS.2020.3047929.
\bibitem{Vaishnav} A. Vaishnav and M. Mandot, Template Matching for Automatic Number Plate Recognition System with Optical Character Recognition, vol. 933. Springer Singapore, 2020. doi: 10.1007/978-981-13-7166-0\_69.
\bibitem{Lu} Q. Lu, Y. Liu, J. Huang, X. Yuan, and Q. Hu, “License plate detection and recognition using hierarchical feature layers from CNN,” Multimed. Tools Appl., vol. 78, no. 11, pp. 15665–15680, 2019, doi: 10.1007/s11042-018-6889-1.
\bibitem{Slimani} I. Slimani, A. Zaarane, W. Al Okaishi, I. Atouf, and A. Hamdoun, “An automated license plate detection and recognition system based on wavelet decomposition and CNN,” Array, vol. 8, no. August, p. 100040, 2020, doi: 10.1016/j.array.2020.100040.
\bibitem{Kessentini} Y. Kessentini, M. D. Besbes, S. Ammar, and A. Chabbouh, “A two-stage deep neural network for multi-norm license plate detection and recognition,” Expert Syst. Appl., vol. 136, pp. 159–170, 2019, doi: 10.1016/j.eswa.2019.06.036.
\bibitem{Ma} L. Ma and Y. Zhang, “Research on vehicle license plate recognition technology based on deep convolutional neural networks,” Microprocess. Microsyst., vol. 82, no. January, p. 103932, 2021, doi: 10.1016/j.micpro.2021.103932.
\bibitem{Burkpalli} V. Burkpalli, A. Joshi, A. B. Warad, and A. Patil, “Automatic Number Plate Recognition Using Tensorflow and Easyocr,” Int. Res. J. Mod. Eng. Technol. Sci., vol. 04, no. 09, pp. 493–501, 2022, [Online]. Available: www.irjmets.com
\bibitem{Abiodun} O. I. Abiodun et al., “Comprehensive Review of Artificial Neural Network Applications to Pattern Recognition,” IEEE Access, vol. 7, pp. 158820–158846, 2019, doi: 10.1109/ACCESS.2019.2945545.
\bibitem{Zhang} X. Y. Zhang, C. L. Liu, and C. Y. Suen, “Towards Robust Pattern Recognition: A Review,” Proc. IEEE, vol. 108, no. 6, pp. 894–922, 2020, doi: 10.1109/JPROC.2020.2989782.
\bibitem{Bernardo} L. S. Bernardo et al., “Handwritten pattern recognition for early Parkinson’s disease diagnosis,” Pattern Recognit. Lett., vol. 125, pp. 78–84, 2019, doi: 10.1016/j.patrec.2019.04.003.
\bibitem{Caggiano} A. Caggiano, J. Zhang, V. Alfieri, F. Caiazzo, R. Gao, and R. Teti, “Machine learning-based image processing for on-line defect recognition in additive manufacturing,” CIRP Ann., vol. 68, no. 1, pp. 451–454, 2019, doi: 10.1016/j.cirp.2019.03.021.
\bibitem{Kattenborn} T. Kattenborn, J. Leitloff, F. Schiefer, and S. Hinz, “Review on Convolutional Neural Networks (CNN) in vegetation remote sensing,” ISPRS J. Photogramm. Remote Sens., vol. 173, no. July 2020, pp. 24–49, 2021, doi: 10.1016/j.isprsjprs.2020.12.010.
\bibitem{Lu} W. Lu, J. Li, J. Wang, and L. Qin, “A CNN-BiLSTM-AM method for stock price prediction,” Neural Comput. Appl., vol. 33, no. 10, pp. 4741–4753, 2021, doi: 10.1007/s00521-020-05532-z.
\bibitem{Tourani}  A. Tourani, A. Shahbahrami, S. Soroori, S. Khazaee, and C. Y. Suen, “A robust deep learning approach for automatic Iranian vehicle license plate detection and recognition for surveillance systems,” IEEE Access, vol. 8, pp. 201317–201330, 2020, doi: 10.1109/ACCESS.2020.3035992.
\bibitem{Izidio} D. M. F. Izidio, A. P. A. Ferreira, H. R. Medeiros, and E. N. d. S. Barros, “An embedded automatic license plate recognition system using deep learning,” Des. Autom. Embed. Syst., vol. 24, no. 1, pp. 23–43, 2020, doi: 10.1007/s10617-019-09230-5.
\bibitem{Mukhija} P. Mukhija, “Wavelet Transform and Morphological Processing Based License Plate Localization,” J. Algebr. Stat., vol. 13, no. 3, pp. 2732–2738, 2022.
\bibitem{Vuola} A. O. Vuola, S. U. Akram, and J. Kannala, “Mask-RCNN and u-net ensembled for nuclei segmentation,” Proc. - Int. Symp. Biomed. Imaging, vol. 2019–April, no. Isbi, pp. 208–212, 2019, doi: 10.1109/ISBI.2019.8759574.
\bibitem{Isa} I. S. Isa, M. S. A. Rosli, U. K. Yusof, M. I. F. Maruzuki, and S. N. Sulaiman, “Optimizing the Hyperparameter Tuning of YOLOv5 for Underwater Detection,” IEEE Access, vol. 10, pp. 52818–52831, 2022, doi: 10.1109/ACCESS.2022.3174583.
\bibitem{Qiao} L. Qiao, Y. Zhu, and H. Zhou, “Diabetic Retinopathy Detection Using Prognosis of Microaneurysm and Early Diagnosis System for Non-Proliferative Diabetic Retinopathy Based on Deep Learning Algorithms,” IEEE Access, vol. 8, pp. 104292–104302, 2020, doi: 10.1109/ACCESS.2020.2993937.
\bibitem{Imran}  A. Imran, J. Li, Y. Pei, J. J. Yang, and Q. Wang, “Comparative Analysis of Vessel Segmentation Techniques in Retinal Images,” IEEE Access, vol. 7, pp. 114862–114887, 2019, doi: 10.1109/ACCESS.2019.2935912.
\bibitem{Zhou} Y. Zhou, S. Bai, C. Wang, X. Chen, E. Fishman, and A. L. Yuille, “Prior-aware Neural Network for Partially-Supervised Multi-Organ Segmentation,” pp. 10672–10681.
\bibitem{Pereira} P. M. M. Pereira et al., “Skin lesion classification enhancement using border-line features – The melanoma vs nevus problem,” Biomed. Signal Process. Control, vol. 57, p. 101765, 2020, doi: 10.1016/j.bspc.2019.101765.
\bibitem{Gao} K. Gao et al., “Dual-branch combination network (DCN): Towards accurate diagnosis and lesion segmentation of COVID-19 using CT images,” Med. Image Anal., vol. 67, p. 101836, 2021, doi: 10.1016/j.media.2020.101836.
\bibitem{LinShao} J. C. W. Lin, Y. Shao, Y. Djenouri, and U. Yun, “ASRNN: A recurrent neural network with an attention model for sequence labeling,” Knowledge-Based Syst., vol. 212, no. xxxx, p. 106548, 2021, doi: 10.1016/j.knosys.2020.106548.
\bibitem{Upadhyay}  U. Upadhyay, F. Mehfuz, A. Mediratta, and A. Aijaz, “Analysis and Architecture for the deployment of Dynamic License Plate Recognition Using YOLO Darknet,” 2019 Int. Conf. Power Electron. Control Autom. ICPECA 2019 - Proc., vol. 2019–November, pp. 1–6, 2019, doi: 10.1109/ICPECA47973.2019.8975456.
\bibitem{XuJiang}  L. Xu, J. Jiang, and L. Liu, “RPNet: A Representation Learning-Based Star Identification Algorithm,” IEEE Access, vol. 7, pp. 92193–92202, 2019, doi: 10.1109/ACCESS.2019.2927684.
\bibitem{Jiao}  S. Jiao, “A CRNN-GRU BASED REINFORCEMENT LEARNING APPROACH TO AUDIO CAPTIONING MoE Key Lab of Artificial Intelligence SpeechLab , Department of Computer Science and Engineering Decoder - 1 La y er 512 Unit GRU,” no. November, pp. 225–229, 2020.
\bibitem{NguyenJatowt} T. T. H. Nguyen, A. Jatowt, M. Coustaty, and A. Doucet, “Survey of Post-OCR Processing Approaches,” ACM Comput. Surv., vol. 54, no. 6, 2021, doi: 10.1145/3453476.
\bibitem{Rais-Zadeh} M. Rais-Zadeh et al., “Gallium nitride as an electromechanical material,” J. Microelectromechanical Syst., vol. 23, no. 6, pp. 1252–1271, 2014, doi: 10.1109/JMEMS.2014.2352617.
\bibitem{Rais-Yang} M. Yang et al., “Symmetry-constrained rectification network for scene text recognition,” Proc. IEEE Int. Conf. Comput. Vis., vol. 2019–October, pp. 9146–9155, 2019, doi: 10.1109/ICCV.2019.00924.
\bibitem{LyuYang} P. Lyu, Z. Yang, X. Leng, X. Wu, R. Li, and X. Shen, “2D Attentional Irregular Scene Text Recognizer,” 2019, [Online]. Available: http://arxiv.org/abs/1906.05708
\bibitem{ZhengWang} Z. Zheng, P. Wang, W. Liu, J. Li, R. Ye, and D. Ren, “Distance-IoU loss: Faster and better learning for bounding box regression,” AAAI 2020 - 34th AAAI Conf. Artif. Intell., no. 2, pp. 12993–13000, 2020, doi: 10.1609/aaai.v34i07.6999.
\bibitem{JayomaMoyon} J. M. Jayoma, E. S. Moyon, and E. M. O. Morales, “OCR Based Document Archiving and Indexing Using PyTesseract: A Record Management System for DSWD Caraga, Philippines,” 2020 IEEE 12th Int. Conf. Humanoid, Nanotechnology, Inf. Technol. Commun. Control. Environ. Manag. HNICEM 2020, 2020, doi: 10.1109/HNICEM51456.2020.9400000.

\end{thebibliography}
\balance

\vfill
\begin{center}

\begin{IEEEbiography}[{\includegraphics*[width=1in]{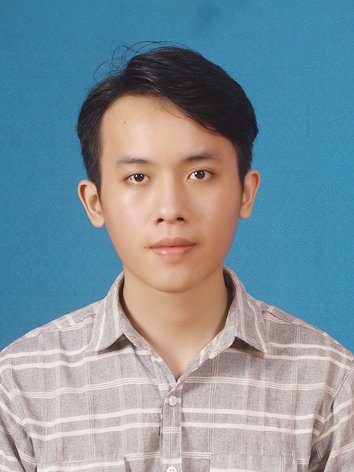}}]
{Dat Tran-Anh} received the B.E. degree in security information and the M.S. degree in computer science from the Posts and Telecommunications Institute of Technology, Hanoi, in 2020 and 2022, respectively. He is currently pursuing the Ph.D. degree in Computer Science at Vietnam Academy of Science and Technology. Since 2023, he has been a lecturer in the Artificial Intelligence department, Thuyloi University. His research interests include image processing and deep learning.
\end{IEEEbiography}

\begin{IEEEbiography}[{\includegraphics*[width=1in]{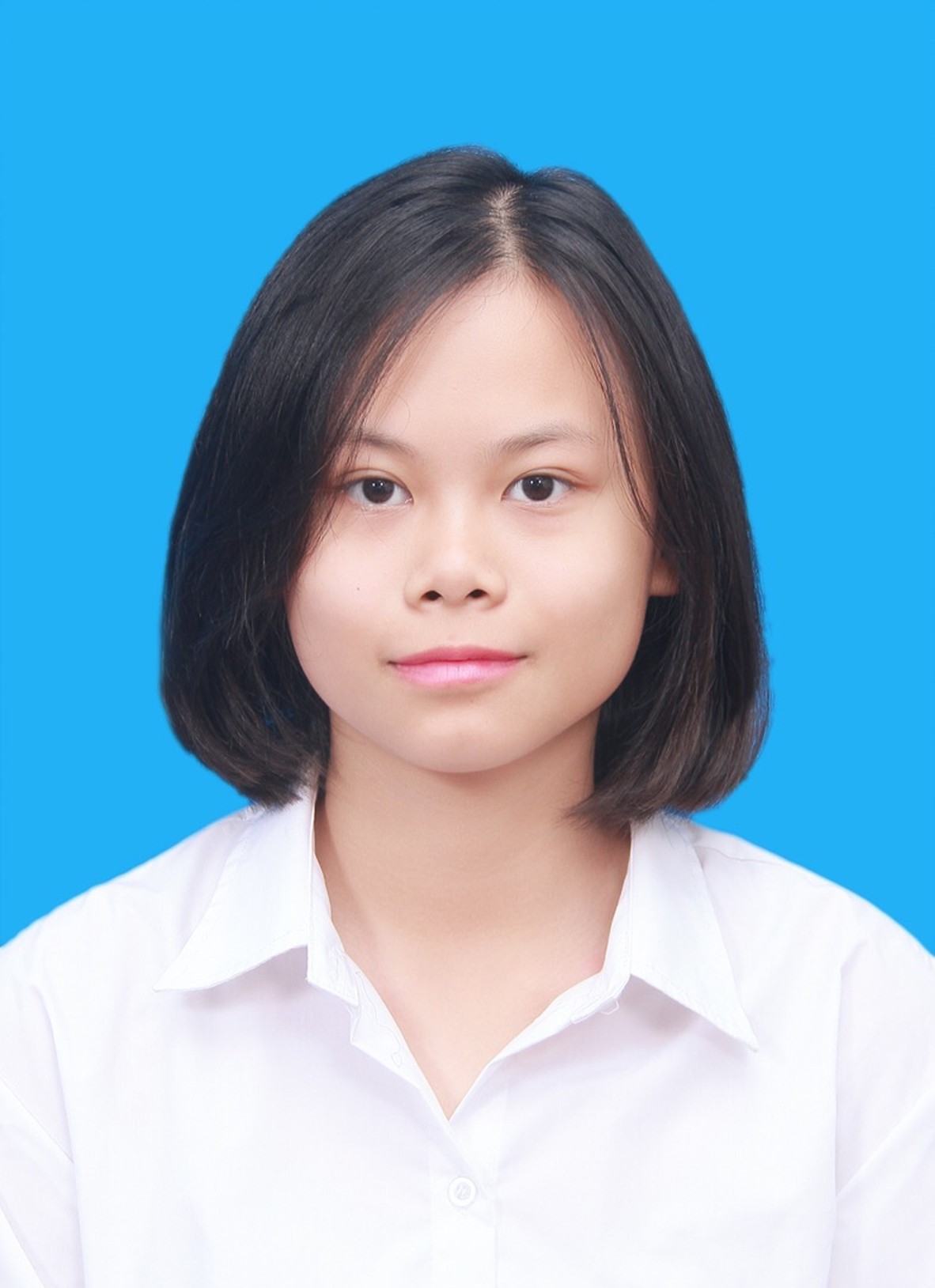}}]
{Khanh Linh Tran} is a third year student at Posts and Telecommunications Institute of Technology majoring in Information Technology. Her research interests are image processing and applications of machine learning/deep learning.
\end{IEEEbiography}

\begin{IEEEbiography}[{\includegraphics*[width=1in]{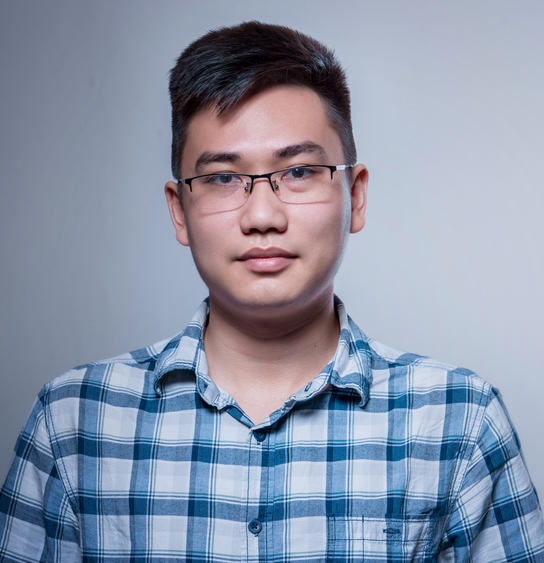}}]
{Hoai-Nam Vu} received the B.E. degrees in Electronic and Telecommunication engineering from the Hanoi University of Science and Technology, Hanoi, Vietnam in 2013 and the M.S. degree in Electronic and Computer engineering from Chonnam National University, Gwangju, South Korea, in 2015. He is currently pursuing the Ph.D. degree in Computer Science at Posts and Telecommunications Institute of Technology, Hanoi. Since 2016, he has been a lecturer with Computer Science Department, Posts and Telecommunications Institute of Technology, Vietnam. His research interests include UAV image processing, machine learning, and deep learning.
\end{IEEEbiography}

\end{center}

\end{document}